\documentclass[twoside,11pt]{article}
\usepackage{jmlr2e}
\usepackage{amsmath}
\usepackage{enumerate}
\usepackage{bbm}

\newtheorem{Theorem}{Theorem}

\newtheorem{Lemma}{Lemma}
\newtheorem{Corollary}{Corollary}
\newtheorem{Proposition}{Proposition}

\newcommand{\field}[1]{\mathbf{#1}}
\newcommand{\Z}{\field{Z}}
\newcommand{\R}{\field{R}}
\newcommand{\E}{\field{E}}

\newcommand{\N}{\field{N}}

\jmlrheading{1}{2013}{1-24}{10/13}{00/00}{Piotr Pokarowski and Jan Mielniczuk}

\ShortHeadings{Combined $\ell_1$ and greedy $\ell_0$ penalized least squares }{Pokarowski and Mielniczuk}

\firstpageno{1}

\begin{document}

\title{ Combined $\ell_1$ and greedy $\ell_0$ penalized least squares \\ for linear model selection}
   \author{\name Piotr Pokarowski \email pokar@mimuw.edu.pl\\
   \addr Faculty of Mathematics, Informatics and Mechanics \\
              University of Warsaw\\
              Banacha 2, 02-097 Warsaw, Poland
   \AND
   \name Jan Mielniczuk \email miel@ipipan.waw.pl \\
   \addr Faculty of Mathematics and Information Science\\
              Warsaw University of Technology\\
              Koszykowa 75,  00-662 Warsaw, Poland
   \AND
   \addr Institute of Computer Science\\
              Polish Academy of Sciences\\
              Jana Kazimierza 5, 01-248 Warsaw, Poland
              }

\editor{}

\maketitle

\begin{abstract}
We introduce a computationally effective algorithm for a linear model selection consisting of three steps:  screening--ordering--selection (SOS). Screening of predictors is based on the thresholded Lasso that is $\ell_1$ penalized least squares.  The screened predictors are then fitted using least squares (LS) and  ordered with  respect to  their $t$ statistics. Finally, a model is selected using greedy generalized information criterion (GIC) that is  $\ell_0$ penalized LS in a nested family induced by the ordering. We give non-asymptotic upper bounds on error probability of each step of the SOS algorithm in terms of both penalties. Then we obtain selection consistency for different ($n$, $p$) scenarios under conditions which are needed for screening consistency of the Lasso. For the traditional setting ($n >p$) we give Sanov-type bounds on the error probabilities of  the ordering--selection algorithm. Its surprising consequence is that the selection error of greedy GIC is asymptotically not larger than of exhaustive GIC. We  also obtain  new bounds on prediction and estimation errors for  the Lasso which are proved in parallel for the algorithm used in practice and its formal version.
\end{abstract}


\section{Introduction}
Literature concerning linear model selection has been lately dominated  by analysis of the {\it least absolute shrinkage and selection operator} (Lasso) that is $\ell_1$ penalized least squares for the 'large $p$ - small $n$ scenario', where $n$ is number of observations and $p$ is number of all predictors. For a broad overview  of the subject we refer to \cite{BuhlmannGeer11}. It is known that  consistency of selection based on the Lasso requires strong regularity of an experimental matrix named  {\it irrepresentable conditions} which are rather unlikely to hold in practice \citep{MeinshausenBuhlmann06, ZhaoYu06}. However, consistency of the Lasso predictors or  consistency of the Lasso estimators of  the linear model parameters  is proved under weaker assumptions such as {\it restricted isometry property} (RIP). The last condition means that singular values of normalized experimental submatrices  corresponding to small sets of predictors are uniformly bounded away from zero and infinity. Under those more realistic conditions and provided that a certain lower bound on the absolute values of model parameters called {\it beta-min condition} holds, the Lasso leads to consistent screening, that is the set of nonzero Lasso coefficients $S$ contains with large predetermined  probability the uniquely defined true model $T$. This property explains  B\"uhlmann's suggestion that one should interpret the second 's' in 'Lasso' as 'screening' rather than 'selection' \citep[see discussion of][]{Tibshirani11} and the task is  now to remove the spurious selected predictors. To this aim two-stage procedures as the adaptive or the thresholded Lasso have been proposed  \citep[cf][]{Zou06,HuangEtAl08,MeinshausenYu09,Zhou09,Zhou10,GeerEtAl11}. They yield selection consistency under strong version of  the beta-min condition  and without such strengthening tend to diminish the number of selected spurious predictors, but, similarly to the Lasso they yield screening consistency only. Alternative approaches require  minimization of more demanding  {\it least squares} (LS) penalized by quasiconvex functions \citep{FanLi01,ZouLi08,ZhangCH10, ZhangT10,ZhangZhang12,HuangZhang12,ZhangT13}. These methods lead to consistent selection under RIP  and considerably weaker version of the beta-min condition, nevertheless are not constructive in that they rely on unknown parameters.

Regularization is required when a matrix is not a full rank or when $n<p$, but for  the traditional regression when an experimental plan is of full rank and  $n>p$ it is possible to construct a computationally effective and selection consistent procedure using  greedy $\ell_0$ penalized  LS, that is a two-stage procedure  {\it ordering--selection} (OS). First,  a full model $F$ using LS is fitted, next predictors are ordered  based on their $t$ statistics from the fit and finally,  a submodel of $F$ in a  nested family pertaining to the ordering  is selected using thresholding as in \cite{RaoWu89} or {\it generalized  information criterion} (GIC) as in \cite{ZhengLoh95}. Sufficient conditions  on an experimental plan and a vector of true coefficients for consistency of such procedures are stated in terms of the {\it Kullback-Leibler divergence} (KL) of the true model from models which do not contain all true predictors  \citep{ZhengLoh95, Shao98, ChenChen08, CasellaEtAl09, PotscherSchneider11, LuoChen13}. In particular, a bound on  the probability of selection error in \cite{Shao98} closely resembles the Sanov theorem in information theory on bounds of  probability of a non-typical event using  the KL divergence.

In our contribution we introduce a computationally effective  three-step algorithm for linear model selection based on a scheme {\it screening--ordering--selection} (SOS). Screening of predictors is based on  a version of the thresholded Lasso proposed by \cite{Zhou09, Zhou10} and yields the screening set $S$ such that $|S|\leq n$. Next, an implementation of OS scheme described above proposed by  \cite{ZhengLoh95} is performed.  We give non-asymptotic upper bounds on error probability of each step of the SOS algorithm in terms of the Lasso and GIC penalties (Theorem 1). As a consequence of proved bounds we obtain  selection consistency for different $(n,p)$ scenarios under  weak conditions which are sufficient for screening consistency of the Lasso. Our assumptions allow for strong correlation between predictors,
in particular replication of spurious predictors is possible.

For case $n>p$ we also give a  bound on probability of selection error of the OS algorithm. Our bound in this case is more general  than in \cite{Shao98} as we allow ordering of predictors, $p=p_n \rightarrow \infty$ and  $|T|=|T_n| \rightarrow \infty$  and moreover GIC penalty may be of order $n$ (Theorem 2). Its simple  but surprising consequence is that the probability of selection error of the greedy GIC is asymptotically not larger than of the exhaustive GIC. Thus employment of greedy search dramatically decreases computational cost of $l_0$ penalized LS  minimization without increasing selection error probability.

As a by-product we obtained a strengthened version of the nonparametric sparse oracle inequality for the Lasso proved by \cite{BickelEtAl09} and, as its consequence, more tight bounds on prediction and estimation error (Theorem 3). We simplified and strengthened  an analogous bound for the thresholded Lasso given by \cite{Zhou09, Zhou10} (Theorem 1 part T1).  It is worth noticing that all results are proved simultaneously for two versions of the algorithm: for the Lasso  used in practice when a response is centered and predictors are standardized as well as for its  formal version for which an intercept corresponds to a dummy predictor.

The paper is organized as follows. In Section 2 the SOS algorithm is introduced and  in Section 3 we  study properties of  geometric characteristics of linear model  pertaining  to an experimental matrix and a vector of coefficients which are related to identifiability of a true model. Section 4 contains our main results that is bounds on selection error probabilities for SOS and OS algorithm. In Section 5 we discuss properties of post-model selection estimators pertaining to SOS. Section 6 contains improved bounds on the Lasso estimation and prediction. Concluding remarks are given in Section 7. Appendix contains detailed  proofs of the results stated in the paper.

\section{Selection algorithm}
The aim of this section is to describe the proposed selection algorithm. As in the first step of the algorithm we use the Lasso estimator to screen predictors and since  in the literature there exist two versions of the Lasso for the linear model  which differ in the treatment of the intercept, we start this section by defining two parametrizations of the linear model related to these versions of the Lasso. Next we state  a general definition encompassing both cases and finally we present our implementation of the SOS scheme.
\subsection{Linear regression model parametrizations}
We consider a general regression model of   real-valued responses having the following structure
\[y_i=\mu(x_{i.}) +\varepsilon_i,\quad\quad i=1,2,\ldots,n,\]
where $\varepsilon_1,\ldots,\varepsilon_n$ are iid $N(0,\sigma^2)$, $x_{i.}\in \R^p$, and $p=p_n$ may depend on $n$. In a vector form we have
\begin{equation}
\label{regmodel}
 y=\mu +\varepsilon,
 \end{equation}
where $\mu=(\mu(x_{1.}),\ldots, \mu(x_{n.}))^{T}, \varepsilon=(\varepsilon_1,\ldots,\varepsilon_n)^{T}$ and $y=(y_1,\ldots,y_n)^{T}$.

Let  $X=[x_{1.},\ldots,x_{n.}]^{T}=[x_1,\ldots,x_p]$ be the $n\times p$ matrix of experiment. We consider two linear parametrizations  of (\ref{regmodel}). The first parametrization is:
\begin{equation}
\label{linregmodel1}
\mu=\alpha^* + X\beta^*,
\end{equation}
where $\alpha^*\in \R$ is an intercept and $\beta^*\in \R^p$ is a vector of coefficients corresponding to predictors. The second parametrization is
\begin{equation}
\label{linregmodel2}
\mu= X\beta^*,
\end{equation}
where the intercept is either set to 0 or is  incorporated  into vector $\beta$ and treated in the same way as all other coefficients in the linear model. In order to treat both parametrizations in the same way  we write $\mu=\tilde X\tilde\beta^*$ where, with ${\mathbbm{1}}_n$ denoting  a column of ones, $\tilde X=[{\mathbbm{1}}_n,X]$ and $\tilde\beta^*=(\alpha^*, {\beta^*}^{T})^{T}$ in the case of (\ref{linregmodel1}) and $\tilde X=X$ and $\tilde \beta^*=\beta^*$ in the case of (\ref{linregmodel2}).

 Let $J\subseteq \{1,2,\ldots,p\}=F$  be an arbitrary subset of the full model $F$  and $|J|$ the number of its elements, $X_J$ is a submatrix of $X$ with columns having indices in $J$, $\beta_J$ is a subvector of $\beta$ with columns having indices in $J$. Moreover,  let $\tilde X_J= [{\mathbbm{1}}_n, X_J]$ and  $\tilde\beta_J=(\alpha,\beta_J^T)^T$ in the case of (\ref{linregmodel1}) or $\tilde X_J= X_J$ and $\tilde\beta_J=\beta_J$  in the case of (\ref{linregmodel2}).  $\tilde H_J$ will stand for  a projection matrix onto the subspace spanned by columns of  $\tilde X_J$. Linear model pertaining to  predictors being columns of $X_J$ will be frequently identified as $J$. We will also denote by $T=T_n$ a  true model that is a model  such that $T={\rm supp}(\beta^*)=\{j \in F: \beta_j^* \neq0\}$ for some $\beta^*$ such that $\mu=\tilde X\tilde\beta^*$. The uniqueness of $T$ and $\beta^*$ for a given $n$ will be discussed in Section~3.
\subsection{Practical and formal Lasso}
The Lasso introduced in \cite{Tibshirani96} is a popular method of estimating $\beta^*$ in  the linear model. For discussion of properties of the Lasso see for example \cite{Tibshirani11} and \cite{BuhlmannGeer11}. When using the Lasso  for data analytic purposes  parametrization (\ref{linregmodel1}) is considered, vector of responses $y$ is centered and columns of $X$ are standardized. The standardization  step is usually omitted in formal analysis in  which  parametrization (\ref{linregmodel2}) is assumed, $\alpha$  is  taken to be 0 and $X$ consists of meaningful predictors only, without column of ones corresponding to intercept \citep[see for example formula 2.1 in ][]{BickelEtAl09}. Here, in order to accomodate both approaches  use a one definition we introduce a general form of the Lasso. Let $H_0$ be  an  $n\times n$  projection matrix, where $H_0$ is specified as  a vector centering matrix
${\mathbbm{I}}_n -{\mathbbm{1}}_n{\mathbbm{1}}_n^{T}/n$ in the case of the applied version of the Lasso pertaining to parametrization (\ref{linregmodel1}) and the identity  matrix ${\mathbbm{I}}_n$ for the formal Lasso corresponding to (\ref{linregmodel2}). Moreover, let
\[ D={\rm diag}(||H_0x_j||)_{j=1}^p,\quad X_0=H_0XD^{-1},\quad X_0=[x_{01},\ldots,x_{0p}],\quad y_0=H_0y\]
and $\theta^*=D\beta^*$, $\mu_0=H_0\mu.$ For estimation of $\beta^*$, data $(X_0,y_0)$ will be used. Note that for the first choice of orthogonal projection in the definition of $X_0$ columns in $X$ are normalized by their norms whereas for the second they are standardized (centered and divided by their standard deviations). Consider the case of (\ref{linregmodel1}) and  denote  by $H_{0J}$ projection onto ${\rm sp}\{(H_0x_j)_{j\in J}\}$. Observe   that as ${\rm sp}\{{\mathbbm{1}}_n,(x_j)_{j\in J}\}={\rm sp}\{{\mathbbm{1}}_n\}\oplus {\rm sp}\{(H_0x_j)_{j\in J}\}$ and consequently
$\tilde H_J=H_{0J} + {\mathbbm{1}}_n{\mathbbm{1}}_n^{T}/n$, we have that
\begin{equation}
\label{projlink}
{\mathbbm{I}}_n - \tilde H_J= ({\mathbbm{I}}_n -H_{0J})H_0.
\end{equation}
The above equality  trivially  holds also in the case of (\ref{linregmodel2}).

For $a=(a_j)\in \R^k$, let $|a|=\sum_{j=1}^k |a_j|$ and $||a||=(\sum_{j=1}^k a_j^2)^{1/2}$ be $\ell_1$ and $\ell_2$ norms, respectively. As $J$ may be viewed as sequence of zeros and ones on $F$, $|J|$ denotes cardinality of $J$.

General form of  the Lasso  estimator of $\beta$ is defined as follows
\begin{equation}
\label{Lasso2}
\hat{\beta}  ={\rm argmin}_{\beta}\{||H_0(y-X\beta)||^2 +2r_L|D\beta|\}=D^{-1}({\rm argmin}_{\theta}\{||y_0-X_0\theta||^2 +2r_L|\theta|\}),
\end{equation}
where a parameter $r_L=r_{nL}$ is a penalty on $l_1$ norm of a potential estimator of $\beta$. Thus in the case of parametrization (\ref{linregmodel1}) the Lasso  estimator of $\beta$ may be defined without using extended matrix $\tilde X$ by  applying $H_0$ to $y-X\beta$ that is by centering it. In the case of  parametrization (\ref{linregmodel2})  $H_0={\mathbbm{I}}_n$ and the usual definition of the Lasso used in  formal analysis is obtained.

Note that in the case of parametrization (\ref{linregmodel1}) $\hat\beta$ is subvector corresponding to $\beta$ of the following minimizer
\begin{equation}
\label{Lasso}
 {\rm argmin}_{\tilde\beta}\{||y-\tilde X\tilde\beta||^2 +2r_L|D\beta|\}= {\rm argmin}_{\alpha,\beta}\{||y-\alpha{\mathbbm{1}}_n - X\beta||^2 +2r_L|D\beta|\},
\end{equation}
where the equality of minimal values of expressions appearing in (\ref{Lasso2})  and (\ref{Lasso})   is obtained when the expression
$||y-\alpha{\mathbbm{1}}_n - X\beta||^2$ is minimized with respect to $\alpha$ for fixed $\beta$. However, omitting centering projection $H_0$ in  (\ref{Lasso2}) when the first column of $X$ consists of ones and corresponds to intercept, leads to lack of invariance of $\hat{\beta}$ when the  data are shifted by a constant and yields different estimates that those used in practice. This is a difference between  the Lasso and the LS estimator:  LS estimator has the same form regardless of  which of the two parametrizations    (\ref{linregmodel1}) or (\ref{linregmodel2}) is applied. Using (\ref{projlink}) we have for  the LS estimator $\hat\beta_J^{LS}$ in model $J$  that the sum of squared residuals for the projection $\tilde Hy$ equals
\begin{equation}
\label{RJ}
R_J=||({\mathbbm{I}}_n -{\tilde H_J})y||^2=||({\mathbbm{I}}_n -H_{0J})y_0||^2=||y_0-X_{0J}\hat\theta_J^{LS}||^2
\end{equation}
and
\[ \hat\beta_J^{LS}=D^{-1}\hat\theta_J^{LS} {\rm,}~~~ \hat\theta_J^{LS}={\rm argmin}_{\theta_J}||y_0-X_{0J}\theta_J||^2.\]
\subsection{Implementation of the screening--ordering--selection scheme}
The SOS algorithm which is the main subject of the paper is the following implementation of the SOS scheme.
\begin{itemize}
\item Screening. Let $\hat\beta=D^{-1}\hat\theta,~~~\hat\theta=(\hat\theta_1,\ldots,\hat\theta_p)^{T}$ be the Lasso estimator with a penalty parameter $r_L$ defined above, and
\[ S_0=\{j:|\hat\theta_j|\geq a_0 \},~~~ S_1=\{j:|\hat\theta_j|\geq  a_1\}, \]
where $a_0=6r_L$ and $a_1=6r_L( |S_0| \vee 1)^{1/2}$.

\item Ordering. Fit the model $ S_1$ using LS and  consider ordering  of predictors $\hat O \equiv \hat O_{ S_1} = (j_1,j_2,\ldots,j_{| S_1|})$ given by decreasing values of  corresponding squared $t$ statistics
$t_{j_1}^2 \geq t_{j_2}^2 \geq \ldots \geq t_{j_{| S_1|}}^2$.

\item Selection. In the nested family  ${\cal G}=\{\emptyset,\{j_1\},\{j_1,j_2\}, \ldots, S_1\}$ choose a model
$\hat T \equiv \hat T_{S_1,\hat O_{S_1}}$ having the smallest value of  the {\it generalized information criterion} (GIC) where $r=r_n$  is a penalty pertaining to GIC and  $R_J$ is defined in (\ref{RJ}). In  the case of multiple minima, the smallest set is taken as $\hat T$.
\end{itemize}

In Section 3 we will discuss conditions under  which $S_1$ includes a unique true model $T$ and $|S_1|\leq n$  with high probability which is necessary for using LS to fit a linear model  in two last steps of the procedure.

We note that empty set  in the definition of ${\cal G}$  corresponds to $\mu=0$ in the case of parametrization (\ref{linregmodel2}) and
$\mu=\alpha$ in the case of (\ref{linregmodel1}).

It is easy to check that
\[ \frac{t_j^2}{n-|S_1|}=\frac{ R_{{S_1}\setminus\{j\}} -R_{S_1} } {R_{{S_1}} }, \]
thus ordering with respect to decreasing values of $(t_j^2)$ in the second step of the procedure is the same as ordering of $(R_{{S_1}\setminus\{j\}}) $ in  decreasing order. We will use this fact in the proof of Lemma \ref{Lem1} in Appendix A.

Computing $(R_J)_{J \in{\cal G}}$  in the selection step demands only one QR decomposition of the matrix $X_{0S_1}$ with columns ordered according to $\hat O$. Indeed, let $X_{0S_1}=QR$, where $Q=[q_1,\dots,q_{|S_1|}]$. The following iterative  procedure can be used
\[R_{\emptyset}=||y_0||^2; \quad {\rm for} \quad k=1,\dots,|S_1|, \quad R_{\{1,\dots,k\}} =R_{\{1,\dots,k-1\}} -(q_k^Ty_0)^2.\]

The OS algorithm is an algorithm intended for the case $p<n$ for which $S_1$ is taken equal to  $F$ and only the two last steps of the SOS are performed.

\section{A true model identifiability}
In this section we consider two  types of linear model characteristics which will be used to quantify the difficulty of selection or, equivalently, a true model identifiability problem, and we study the interplay between them.
\subsection{Kullback-Leibler divergences}
Let $T$ be given true model that is $T\subseteq F$ such that $\mu=\tilde X\tilde\beta^*=\tilde X_T\tilde\beta_T^* $ and
$T={\rm supp}(\beta_T^*)=\{j \in F: \beta_{j,T}^* \neq0\}$. For  $J\subseteq F$ define
\[ \delta(T\parallel J)=||(\mathbbm{I}_n-\tilde{H}_J)\tilde X_T{\tilde{\beta^*_T}}||^2. \]
In view of (\ref{projlink}) we obtain
\begin{equation}
\label{delta}
\delta(T\parallel J)=||(\mathbbm{I}_n-H_{0J})H_0\tilde X_T{\tilde{\beta^*_T}}||^2=||(\mathbbm{I}-H_{0J})H_0 X_T{{\beta}^*_T}||^2=||(\mathbbm{I}-H_{0J}) X_{0T}{{\theta}^*_T}||^2.
\end{equation}

Let $KL({\tilde\beta^*_T}\parallel {\tilde\beta_J})=\E_{\tilde\beta^*_T}\log (f_{\tilde\beta^*_T}/f_{\tilde\beta_J})$ be the {\it Kullback-Leibler divergence} of normal density  $f_{\tilde\beta_T^*}$ of $N(\tilde X_{T}{\tilde\beta^*_T},\sigma^2\mathbbm{I}_n)$ from the normal density  $f_{\tilde\beta_J}$ of $N(\tilde X_{ J}{\tilde\beta_J},\sigma^2\mathbbm{I}_n).$ Let $\Sigma=X_0^{T}X_0$  be a coherence matrix if  $H_0$ is the identity matrix and a correlation matrix if  $H_0={\mathbbm{I}}_n -{\mathbbm{1}}_n{\mathbbm{1}_n^{T}}/n$. $\Sigma_J$ stands for  a submatrix of $\Sigma$ with columns having indices in $J$ and its eigenvalues will be called {\it sparse eigenvalues} of $\Sigma$. In particular, $\lambda_{min}(\Sigma_{J})$ denotes the smallest eigenvalue of $\Sigma_{J}$. The following proposition  lists the basic properties of the parameter $\delta$. Observe also that $\delta(T\parallel J)$ is a parameter of noncentrality of $\chi^2$  distribution of $R_J$ that is $R_J\sim \chi_{n-|J|}^2(\delta(T\parallel J))$.

\begin{Proposition}
\label{Prop1}
\begin{equation}\nonumber
\label{deltaProp} (i)
\quad\delta(T\parallel J)=2\sigma^2 \min_{\tilde\beta_J} KL({\tilde\beta^*_T}\parallel {\tilde\beta_J}) = 2\sigma^2 \min_{\tilde\beta_J} KL({\tilde\beta_J}\parallel {\tilde\beta^*_T}).
\end{equation}
\begin{equation}
\label{deltaLambda} (ii)
\quad \delta(T \parallel J)=\min_{\theta_J}\Bigg|\Bigg| [X_{0,T\setminus J},X_{0,J}]
\begin{pmatrix}  \theta^*_{T\setminus J} \cr
 \theta_{J}  \cr
\end{pmatrix}\Bigg|\Bigg|^2 \geq \lambda_{min}(\Sigma_{J\cup T})||\theta_{T\setminus J}^*||^2
\end{equation}
\end{Proposition}

The following scaled Kullback-Leibler divergence will be employed in our main results in Section 4.
\begin{equation*}
\delta(T,s)=\min_{j\in T,J\supseteq T,|J|\leq s}\delta(T \parallel J\setminus\{j\}).
\end{equation*}

This coefficient was previously used to prove selection consistency  in \cite{ZhengLoh95, ChenChen08, LuoChen13} and to establish asymptotic law of post-selection estimators in \cite{PotscherSchneider11}. Similar coefficients appear to prove selection consistency in \cite{Shao98} and \cite{CasellaEtAl09}. Obviously, $\delta(T,s)$ is a nonincreasing function of $s$.

Identifiability of a true model is stated in the proposition below in terms of
\[ \delta(T)=\min_{J\nsupseteq T, |J|\leq |T|}\delta(T \parallel J). \]
\begin{Proposition}
\label{deltaUnique}
There exists at  most one true model $T$ such that $\delta(T)>0$.
\end{Proposition}
Assume by contradiction that $T'$ is a different true model,  that is we have $T'={\rm supp}(\beta)$ for some $\tilde\beta$ such that $\mu=\tilde X\tilde\beta$. Then by symmetry we can assume $|T| \leq |T'|$. Hence $|T'\setminus T| >0$ and $\delta(T') \leq \delta(T' \parallel T)=0$. It is easy to see that if $\delta(T)>0$  then columns of $X_T$ are linearly independent and, consequently, there exists at most one $\tilde\beta_T^*$ such that $\mu=\tilde X_T \tilde\beta_T^*$.

In Section 4.2 we infer identifiability of a true model $T$  from Proposition~\ref{deltaUnique} and the following inequality
\begin{equation}
\label{deltaDelta}
\delta(T,p) \leq \delta(T).
\end{equation}
Indeed, for any $J$ such that $J\nsupseteq T$ and $|J|\leq |T|$ there exists $j\in T$ such that $J \subseteq F\setminus{\{j\}}$.
Thus we obtain $\delta(T \parallel F\setminus\{j\}) \leq \delta(T \parallel J)$ and minimizing both sides yields (\ref{deltaDelta}).
\subsection{Restricted eigenvalues}
For $J\subseteq F$, ${\bar J}= F\setminus{ J}$ and $c>0$ let
\begin{equation*}
\kappa^2(J,c)=\min_{\nu \neq 0, |\nu_{\bar J}|\leq c|\nu_{J}|} \frac{\nu^{T}\Sigma \nu}{\nu_J^{T}\nu_J} \quad\quad {\rm and} \quad\quad
\kappa^2(s,c)=\min_{J:|J|\leq s}\kappa(J,c).
\end{equation*}
Both coefficients will be called {\it restricted eigenvalues} of $\Sigma$. Observe that
\begin{equation}
\label{kappaGamma}
\kappa^2(J,c)= \min_{\nu \neq 0, |\nu_{\bar J}|\leq c|\nu_{J}|} \frac{||X_0\nu||^2}{||\nu_J||^2}
               = \min_{\nu \neq 0, |\nu_{\bar J}|\leq c|\nu_{J}|}\frac{||X_0\nu_{J}- X_0\nu_{\bar J}||^2}{||\nu_J||^2}.
\end{equation}

The coefficient $\kappa(s,c)$ is a modified version of an index  introduced in \cite{BickelEtAl09}. Modification consists in replacing $X$ appearing in the original definition by $X_0$ and omitting the term $n^{-1/2}$. Pertaining parameters  for  a fixed set of predictors $J$ and their various modifications were introduced and applied to bound the Lasso errors by \cite{GeerBuhlmann09}.

In order to study relations between  sparse and restricted eigenvalues we set
\begin{equation*}
\kappa^2(J,0)=\min_{\nu \neq 0, {\rm supp}(\nu)\subseteq J}\frac{\nu^{T}\Sigma \nu}{\nu^{T}\nu}
  \quad\quad {\rm and} \quad\quad \kappa^2(s,0)= \min_{J:|J|\leq s} \kappa^2(J,0).
\end{equation*}
From Rayleigh-Ritz theorem we have
\begin{equation}
\label{kappa0}
\kappa^2(J,0) = \lambda_{min}(\Sigma_{J}) \leq \frac{tr(\Sigma_J)}{|J|}= 1.
\end{equation}
Note that $\kappa(J,c)$ and  $\kappa(s,c)$ are nonincreasing functions of both arguments.  Moreover, $\kappa^2(J,c)\leq \kappa^2(J,0)$ and
$\kappa^2(s,c)\leq \kappa^2(s,0)$. This holds in  view of an observation that  for any fixed $J$ and $c>0$, any $\nu$ such that  ${\rm supp} (\nu) \subseteq J$ satisfies  $\nu=\nu_J$ and thus $|\nu_{\bar J}|\leq c|\nu_{J}|$. It is easy to show also that $\kappa^2(J,c)\to \kappa^2(J,0)$ and $\kappa^2(s,c)\to \kappa^2(s,0)$  monotonically when $c\to 0^+$.  Another less obvious bound, which  is used in the following is stated below.
\begin{Proposition}
\label{kappaProp}
For any $s\in \N$ and $c>0$
\[\kappa^2(s,c)\leq (\lfloor c\rfloor+1)\kappa^2(\lfloor c\rfloor+1)s,0).\]
\end{Proposition}

Condition $\kappa(s,c)>0$ imposed on matrix $X$ is called {\it restricted eigenvalue condition} in \cite{BickelEtAl09} for their slightly different $\kappa$. Proposition \ref{kappaProp} generalizes an observation there (p.1720) that  if the restricted eigenvalue condition holds for $c \geq 1$,  then all  square submatrices of $\Sigma$ of size $2s$ are necessarily positive definite. Indeed, the proposition above implies that $\kappa(2s,0)>0$ from which the observation follows. Positiveness of $\kappa(T,c)$ which due to the restriction on vectors  $\nu$ over which minimization is performed can hold even for $p>n$, is a certain condition on weak correlation of columns. This condition, which will be assumed later, is much less stringent than $\kappa(|T|,c)>0$, as it allows for example replication of columns belonging to the complement of $T$. Moreover $\kappa(T,c)>0$ for $c \geq 1$ implies identifiability of a true model.
\begin{Proposition}
\label{kappaUnique}
There exists at  most one true model $T$ such that $\kappa(T,1)>0$.
\end{Proposition}

It follows that if $\kappa(T,1)>0$,  then columns of $X_T$ are linearly independent and, consequently, there exists at most one $\tilde\beta_T^*$ such that $\mu=\tilde X_T \tilde\beta_T^*$.

The following $\kappa-\delta$ {\it inequalities} follow from the  Propositions~\ref{Prop1} (ii) and the Proposition~\ref{kappaProp}.
We set $\theta^*_{\min}=\min_{j\in T}|\theta^*_j|$ and $t=|T|$.

\begin{Proposition} We have
\label{kappaDelta}
\begin{equation}
\label{betadelta1}
\kappa^2(T,3)\theta^{*2}_{\min} \leq \delta(T,t)
\end{equation}
and
\begin{equation}
\label{betadelta2}
\kappa^2(t,3)\theta^{*2}_{\min} \leq 4\delta(T,4t).
\end{equation}
\end{Proposition}

\section{Error bounds for SOS and OS algorithms}
In this section we present the main result that is nonasymptotic bounds on the error probabilities for all steps of the algorithm SOS. The errors of consecutive steps of the SOS  constitute decomposition of  the selection error  into four parts. Two errors which can be possibly  committed in the selection step correspond to two situations when the selected model is a proper subset or  a superset of $T$.
\subsection{Error bounds for  SOS}
Let ${\cal S}_n$ be a family of  models having no more than $s$ predictors and ${\cal T}_n=\{S \in {\cal S}_n : S\supseteq T\}$ consists of all true models in ${\cal S}_n$. Observe that $|{\cal T}_n|=\sum_{k=0}^{s-t}{p-t\choose k}$. Moreover, let $O_{S_1}$ denote a set of all correct orderings of $S_1$ that is orderings such that all true variables in $S_1$ precede the spurious ones. To simplify notation set $\delta_s=\delta(T,s)$,  $\delta_t=\delta(T,t)$ and  $\kappa=\kappa(T,3)$. We also  define two constants $c_1=(3+6\sqrt{2})^{-1}\approx0.087$ and $c_2=(6+4\sqrt{2})^{-1}\approx0.086$. We assume for the remaining part of the paper that $p\geq t+1 \geq 2$ as boundary cases are easy to analyse. Moreover, we assume the following condition which ensures that the size of $S_1$ defined in the first step of SOS algorithm does not exceed $n$ with large probability and consequently  LS could be performed on data $(y_0,X_{0S_1})$. It states that
 \begin{equation}
\label{weakcorrelation}
s=s(T)= t +\lfloor t^{1/2}\kappa^{-2}\rfloor \leq  n.
\end{equation}

\begin{Theorem} \label{Th1}
(T1) If for some $a\in(0,1)$ $8a^{-1}\sigma^2\log p  \leq  r_L^2\leq c_1^2t^{-1}\kappa^4\theta_{min}^{*2}$, then
\begin{equation}
\label{L21}
P( S_1\not\in {\cal T}_n)  \leq
\exp \bigg(-\frac{(1-a)r_L^2}{8\sigma^2}\bigg)\bigg(\frac{\pi r_L^2}{8\sigma^2}\bigg)^{-1/2}.
\end{equation}
(T2) If for some $a\in(0,1)$ $a^{-1}\sigma^2\log p  \leq  c_2(s-t+2)^{-1}\delta_s$, then
\begin{equation}
\label{L22}
P( S_1\in {\cal T}_n, \hat O\not \in O_{S_1} ) \leq
\frac{3}{2} \exp \bigg(-\frac{(1-a)c_2\delta_s}{\sigma^2}\bigg) \bigg(\frac{\pi c_2\delta_s}{\sigma^2}\bigg)^{-1/2}.
\end{equation}
(T3) If for some $a\in (0,1)$   (a) $r < at^{-1}\delta_t$ and  (b) ${8a^{-1}\sigma^2}\log t \leq {(1-a)^2\delta_t}$, then
\begin{equation}
\label{L23}
P( S_1\in {\cal T}_n, \hat O\in O_{S_1}, |\hat T|<t ) \leq
\frac{1}{2} \exp \bigg(-\frac{(1-a)^3\delta_t}{8\sigma^2}\bigg)\bigg(\frac{\pi (1-a)^2\delta_t}{8\sigma^2}\bigg)^{-1/2}.
\end{equation}
(T4) If for some $a\in (0,1)$ $4a^{-1}\sigma^2\log p\leq r$, then
\begin{equation}
\label{L24}
P( S_1\in {\cal T}_n, \hat O \in O_{S_1}, |\hat T |>t ) \leq
\exp \bigg(-\frac{(1-a)r}{2\sigma^2}\bigg)\bigg(\frac{\pi r}{2\sigma^2}\bigg)^{-1/2}.
\end{equation}
\end{Theorem}

A regularity condition on the plan of experiment induced by the assumption of Theorem~\ref{Th1} (T1),  namely
$8a^{-1}\sigma^2\log p\leq c_1^2t^{-1}\kappa^4\theta_{min}^{*2}$, is known as {\it beta-min condition}. Its equivalent form, which is  popular in the literature states  that for some $a \in (0,1)$
\begin{equation}
\label{betamin}
\sqrt{8c_1^{-2}a^{-1}\sigma^2t\kappa^{-4}\log p} \leq  \min_{j\in T}||H_0x_j||\,|\beta_j^*|.
\end{equation}
Observe that (\ref{betamin}) implies that $\kappa=\kappa(T,3)>0$, so it guarantees identifiability of $T$ in view of Proposition \ref{kappaUnique}.

Note that  bounds in (T2) and (T3) as well as the bounds in Theorem 2 below can be interpreted as results analogous to the Sanov theorem in information theory on bounding probability of a non-typical event (c.f. for example \cite{CoverThomas06}, Section 11.4), as in view of Proposition \ref{Prop1} (i) $\delta_s$ may be expressed as $\min_{\beta \in B} 2\sigma^2 KL(\beta \parallel \beta^*)$ for a certain set $B$.

The first corollary provides an upper bound on a selection error of the SOS algorithm under simpler conditions. The assumption $r_L^2= 4r$ is quite arbitrary, but results in the same lower bound for penalty and almost the same bound on error probability as in the Corollary \ref{cor3} below. Note that boundary values of $r_L^2$ and $r$ of order $\log p$ are allowed in Corollaries \ref{cor1}--\ref{cor3}.
\begin{Corollary}
\label{cor1}
If $r_L^2= 4r$ and for some $a\in (0,1-c_1)$ we have\\
(i) $4a^{-1}\sigma^2\log p \leq r \leq (c_1^2/4)at^{-1}\kappa^4\theta_{min}^{*2}$ and
(ii) $r \leq (4c_2/3)t^{-1/2}\kappa^2\delta_s$, then
\begin{equation*}
 P(\hat T\neq T)  \leq  4 \exp \bigg(-\frac{(1-a)r}{2\sigma^2}\bigg) \bigg(\frac{\pi r}{2\sigma^2}\bigg)^{-1/2}.
 \end{equation*}
\end{Corollary}

We consider now the results above under   stronger conditions. We replace $\kappa$ in  the assumption (\ref{betamin}) by
smaller $\kappa_{min}=\kappa(t,3)$ and we assume the following {\it weak correlation condition}
\begin{equation}
\label{weak2}
\kappa_{min}^{-2}\leq 3t^{1/2},
\end{equation}
which is weaker than a condition $\kappa_{min}^{-2}\leq t^{1/2}$ in Theorem 1.1 in \cite{Zhou09, Zhou10}. The condition (\ref{weak2}) implies in view of (\ref{weakcorrelation}) that the size of the screening set $S_1$ satisfies $s\leq 4t$. Thus we obtain from (\ref{betadelta2}) that
$ (c_1^2/4)at^{-1}\kappa_{min}^4\theta_{min}^{*2} < (4c_2/3)t^{-1/2}\kappa_{min}^2\delta_s$ as $\delta_s\geq \delta_{4t}$ and $16c_2/(3c_1^2)\geq 1$. Hence the Corollary \ref{cor1} simplifies to the following corollary.
\begin{Corollary}
\label{cor2}
Assume  (\ref{weak2}) and $r_L^2=4r$. If for some $a\in (0,1-c_1)$ \\
$4a^{-1}\sigma^2\log p \leq r \leq (c_1^2/4)at^{-1}\kappa_{min}^4\theta_{min}^{*2}$,
then the conclusion of Corollary \ref{cor1} holds.
\end{Corollary}

Theorem \ref{Th1}  shows that the SOS algorithm is an improvement of the adaptive and the thresholded Lasso \citep[see][]{Zou06,HuangEtAl08,MeinshausenYu09,Zhou09,Zhou10,GeerEtAl11} as under weaker assumptions on an experimental matrix  than assumed there we obtain much stronger result, namely selection consistency. Indeed,  assumptions of Theorem \ref{Th1} are stated in terms of $\kappa(T,3)$, $\delta_s$ and $\delta_t$ instead of $\kappa(t,3)$, thus allowing for example replication of spurious predictors. Discussion of assumptions of Corollary 2 shows that the original conditions in  \cite{Zhou09,Zhou10} are stronger than our conditions ensuring screening consistency of  the thresholded Lasso. We stress also that the bound of Theorem 2 are valid in both cases when the formal or the practical Lasso is used in the screening step.

The SOS algorithm also turns out to be  a competitor of iterative approaches which require  minimization of more demanding  LS penalized by quasiconvex functions \citep{FanLi01, ZouLi08, ZhangCH10, ZhangT10, ZhangZhang12, HuangZhang12, ZhangT13}.  Consider {\it multistage convex relaxation} (MCR)  studied in \cite{ZhangT10,ZhangT13} which  is the latest example of this group of algorithms. Though the method is defined without any normalization of X, the condition on correlation of predictors  assumed there  seems to be stronger than ours. It is similar to RIP and assures that for some $a \in (0,1)$ the condition $1-a \leq ||X\nu||^2/||\nu||^2 \leq 1+a$ holds for all $\nu$ such that  ${\rm supp}(\nu)\leq 6t+1$. Moreover, our algorithm depends on penalties  $r$ and $r_L$ whereas  MCR also involves a minimal sparse eigenvalue $\rho_-=\min_{\nu : {\rm supp}(\nu)\leq 4t}  ||X\nu||^2/||\nu||^2$. On the other hand, the beta-min condition assumed in \cite{ZhangT13} is:
\[ \sqrt{const\sigma^2\rho_-^{-2}\log p} \leq \min_{j \in T}|\beta_j^*| .\]
The comparison of  $\rho_-$ with our $t^{-1/2}\kappa^2(T,3)$ remains  an open problem.
\subsection{Error bounds for OS}
Now we state the corresponding bounds for error probabilities of OS algorithm in the case of $p\leq n$. We recall that in the case of OS  $S_1=F$. Thus ${\cal S}_n={\cal T}_n=\{S_1\}$ and $P( S_1\not\in {\cal T}_n)=0$.

\begin{Theorem}
\label{Th2}
If for some $a\in (0,1)$ $~~a^{-1}\sigma^2\log(t(p-t)) \leq c_2\delta_p$, then
\begin{equation*}
\label{T2+}
P( \hat O\not \in O ) \leq
\frac{3}{2} \exp\bigg( -\frac{(1-a)c_2\delta_p}{\sigma^2}\bigg) \bigg( \frac{\pi c_2\delta_p}{\sigma^2}\bigg)^{-1/2}.
\end{equation*}
Moreover, (T3) and (T4) of Theorem \ref{Th1} hold.
\end{Theorem}

Observe that assumptions of Theorem \ref{Th2} imply that $\delta_p>0$ which  guarantees uniqueness of $T$ in view of (\ref{deltaDelta}).

The next corollary is analogous to Corollary \ref{cor1} and provides an upper bound for a selection error of OS algorithm under simpler conditions.
This bound is more general  than in \cite{Shao98} as we allow greedy selection (specifically ordering of predictors), $p=p_n \rightarrow \infty$, $t=t_n \rightarrow \infty$  and moreover GIC penalty may be of order $n$.

\begin{Corollary}
\label{cor3}
If for some $a\in (0,2c_2)$ $~4a^{-1}\sigma^2\log p \leq r \leq \min \big( at^{-1}\delta_t,~2c_2\delta_p \big)$, then
\begin{equation*}
 P(\hat T_{OS}\neq T)  \leq
3 \exp \bigg(-\frac{(1-a)r}{2\sigma^2}\bigg)\bigg(\frac{\pi r}{2\sigma^2}\bigg)^{-1/2}.
\end{equation*}
\end{Corollary}

It is  somewhat surprising consequence of the Corollary \ref{cor3} that, from an asymptotic point of view, the selection error of the OS algorithm, which is a version of the greedy GIC, is not greater than the selection error of the plain, exhaustive GIC. Specifically, if we define the exhaustive GIC selector by
\[{\hat T}_E={\rm argmin}_{J: J\subseteq F, |J|\leq n}\{R_J+|J|r\},\]
then it follows from the lower bound in (\ref{boundschi1}) below, that  for an arbitrary fixed index $j_0\not\in T$ and $r>0$ we have
\begin{equation}
\label{errE}
 P({\hat T}_E\neq T)\geq P(R_{T\cup \{j_0\}}-R_T>r)\geq
\frac{r}{r+\sigma^2} \exp \bigg(-\frac{r}{2\sigma^2} \bigg) \bigg(\frac{\pi r}{2\sigma^2}\bigg)^{-1/2}.
\end{equation}
If the penalty term satisfies $\log p \ll r \ll \min(\delta_t/t,~\delta_p)$ for $n \rightarrow \infty$, then from Corollary \ref{cor3} and (\ref{errE}) we obtain
\begin{equation}
\label{greedGood}
\varlimsup_n \log P(\hat T_{OS}\neq T) \leq  \varliminf_n \log P({\hat T}_E\neq T).
\end{equation}
The last inequality indicates that it pays off to apply the greedy algorithm in this context as  a greedy search dramatically reduces $\ell_0$ penalized LS without increasing its selection error.

The bounds on the selection error given in Corollaries \ref{cor1}--\ref{cor3} imply consistency of the SOS and OS provided $r_n \rightarrow \infty$ and its strong consistency provided $r_n\geq c\log n$ for some $c>2\sigma^2/(1-a)$. For boundary penalty $r_n=4a^{-1}\sigma^2\log p_n$ where $a \in (0,2c_2)$, we obtain strong consistency of these algorithms if $n^{ca/(1-a)}\leq p_n$ for some  $c >0.5$. Comparison of selection errors probabilities of the SOS and OS algorithms for $p<n$ requires further research.

\section{Properties of post-model selection estimators}
We list now several properties of   post-model selection estimators which follow from the main results.
Let $\hat {\cal B}={\cal B}(\hat T,y)$ be any event defined in terms of given selector $\hat T$  and $y$ and ${\cal B}={\cal B}(T,y)$ be an analogous event pertaining to $T$ and $y$. Let ${\cal B}^c$ and ${\hat {\cal B}}^c$ be complements of ${\cal B}$ and ${\hat {\cal B}}$, respectively. Observe that we have
\[P(\hat {\cal B})\leq P(\hat {\cal B},\hat T\neq T) +P(\hat T\neq T)\leq P({\cal B}) + P(\hat T\neq T).\]
Analogously, $P({\hat {\cal B}}^c) \leq P({\cal B}^c) + P(\hat T\neq T)$, which implies $P({\cal B})\leq P(\hat {\cal B}) + P(\hat T\neq T)$.
Both inequalities yield
\begin{equation}
\label{postSelEstim}
 | P(\hat {\cal B})-P({\cal B})|\leq P(\hat T\neq T)
\end{equation}

In particular, when ${\cal B}=\{G >u\}$ and $\hat {\cal B}=\{\hat G>u\}$ and  $G$ is some  pivotal quantity
then  (\ref{postSelEstim}) implies  that $P(\hat {\cal B})$ is approximated by  $P({\cal B})$ uniformly in $u$. For example,  let $\hat{\tilde\beta}_T$  denote the LS estimator fitted on $T$,
 $d=t+1$ for parametrization~(\ref{linregmodel1}) and $d=t$ for parametrization~(\ref{linregmodel2}) and define
\[  f=f(T,y)= \frac{||\tilde X_T\hat{\tilde\beta}_{T}^{LS} - \tilde X_T\tilde\beta_T^*||^2/d} {||y-\tilde X_T\tilde\beta_{T}^{LS}||^2/(n-d)}.  \]
Observe that the variable $f$ follows a Fisher-Snedecor  distribution  ${\cal F}_{d,n-d}$.
Then the bound on the selection error given in Corollary \ref{cor1}, the assumption  $\varepsilon \sim N(0,\sigma^2{\mathbbm{I}}_n)$ and (\ref{postSelEstim}) imply the following corollary.
\begin{Corollary}
 Assume that  conditions of Corollary 1 are  satisfied. Then
 \begin{equation*}
\sup_{u\in R} | P(\hat f \leq u) -P(f \leq u) |\leq  4 \exp \bigg(-\frac{(1-a)r}{2\sigma^2}\bigg)\bigg(\frac{\pi r}{2\sigma^2}\bigg)^{-1/2}.
 \end{equation*}
 \end{Corollary}
Note that any apriori upper bound on $d$ in conjunction with Corollary 4 yields an approximate confidence region for $\tilde \beta_{\hat T}^*$.

Moreover, it follows from the Corollary \ref{C3} below that  the Lasso estimator has  the following estimation and prediction errors
\begin{Corollary}
Assume that  conditions of Corollary \ref{C3} are  satisfied. Then
\begin{equation*}
||X\hat \beta - X\beta^*|| = O_P\big(t_n^{1/2}\kappa_n^{-1}\sqrt{\log p_n}\big), \quad\quad
|D(\hat\beta-\beta^*)| = O_P\big(t_n\kappa_n^{-2}\sqrt{\log p_n}\big).
\end{equation*}
\end{Corollary}
Analogous properties of post-selection  estimator  of $\beta^*$ are given below. 
\begin{Corollary}
 Assume that  conditions of Corollary 1 are  satisfied. Then
 \begin{equation*}
||X\hat \beta_{SOS} - X\beta^*|| = O_P\big(t_n^{1/2}\big), \quad\quad
|D(\hat\beta_{SOS}-\beta^*)| = O_P\big(t_n\lambda_n^{-1/2}\big),
\end{equation*}
where $\lambda_n=\lambda_{min}(\Sigma_T)$.
\end{Corollary}
In view of the inequality $\kappa_n^2<\lambda_n$ it is seen that the estimation and prediction rates for the SOS post-selection estimator given
above  are better by the factor $\kappa_n^{-1}\sqrt{\log p_n}$ than the corresponding rates for the Lasso.

\section{Error bounds for the Lasso estimator}
We assume from now on that   the general model (\ref{regmodel}) holds.
Let $\mu_0=H_0\mu$, $\mu_{\beta}=H_0X\beta=X_0\theta$ for an arbitrary $\beta\in \R^p$ and $\mu_{\hat\beta}=H_0X\hat\beta=X_0\hat\theta$. Moreover, $\Delta=\hat\theta-\theta=D(\hat\beta-\beta)$ and  recall that $\Delta_J$ stands for subvector of $\Delta$ restricted to coordinates in $J$ and $J_a={\rm supp}(a) = \{j:a_j\neq 0\}$. Finally let ${\cal A}=\bigcap_{j=1}^p\{2|x_{0j}^{T}\varepsilon|\leq r_L\}$ and ${\cal A}^c$ be a complement of ${\cal A}$.  From the Mill inequality (see the  right hand side  inequality in (\ref{boundschi1}) below) we obtain for $Z\sim N(0,1)$
\begin{equation}
\label{boundA}
P({\cal A}^c)\leq \sum_{j=1}^{p} P(2|x_{0j}^T\varepsilon|>r_L) = pP\big(Z^2 > \frac{r_L^2}{4\sigma^2}\big)\leq
p \exp \bigg(-\frac{r_L^2}{8\sigma^2}\bigg)\bigg(\frac{\pi r_L^2}{8\sigma^2}\bigg)^{-1/2}.
\end{equation}
As a by-product of the proofs of  the theorems above we state in this section a strengthened version of the Lasso error bounds and their consequences.
\begin{Theorem}
\label{SOI}
(i) On ${\cal A}$ we have
\begin{equation}
\label{eq3}
||\mu_0- \mu_{\hat\beta}|| \leq ||\mu_0-\mu_{\beta}||+ 3r_L|J_\beta|^{1/2}\kappa^{-1}(J_\beta,3).
\end{equation}
(ii) Moreover, on the set ${\cal A}\cap\{\beta:\,|\Delta|\leq 4|\Delta_J|\}$ we have
\begin{equation}
\label{eq3+}
r_L|\Delta|\leq 2||\mu_0- \mu_{\beta}||^2 +{8r_L^2|J_\beta|}{\kappa^{-2}}(J_\beta,3).
\end{equation}
\end{Theorem}

Squaring both sides of (\ref{eq3}) yields the following bound
\begin{equation*}
\label{bogus}
||\mu_0- \mu_{\hat\beta}||^2 \leq (||\mu_0-\mu_{\beta}||+ \frac{3r_L|J_\beta|^{1/2}}{\kappa(J_\beta,3)})^2=\inf_{a>0}(1+a)\Big(||\mu_0- \mu_{\beta}||^2 +\frac{9r_L^2|J_\beta|}{a\kappa^2(J_\beta,3)}\Big),
\end{equation*}
where the equality above is easily seen. Obviously $\kappa(|J_\beta|,3) \leq \kappa(J_\beta,3)$, hence (\ref{eq3}) is tighter than Theorem 6.1 in  \cite{BickelEtAl09} (we disregard  a small difference in normalization of $X$ mentioned in Section 3). Moreover, the  bound  above is valid  for  both the practical and the formal Lasso.

Let us note that as $\beta$ in (\ref{eq3}) is arbitrary,  the minimum  over all $\beta\in \R^P$ can be taken. Analogously we can minimize the right hand side of (\ref{eq3+}) over all $\beta:\,|\Delta|\leq 4|\Delta_J|$. Note also that if a parametric model $\mu=\tilde X_J \tilde \beta_J$ holds, then  (\ref{eq1}) below   implies that indeed  a condition  $|\Delta|\leq 4|\Delta_J| $  is satisfied.  The next corollary strengthens the $\ell_1$ estimation error inequality (7.7) and the predictive inequality (7.8) in Theorem 7.2 in \cite{BickelEtAl09}. Note that $X$ below  does not need to have normalized columns and the constant appearing in (7.7) and (7.8) in \cite{BickelEtAl09} is 16.
\begin{Corollary} Let $\beta$ be such that $\mu_0=\mu_\beta$. Then  (\ref{eq3+}) and (\ref{eq3}) have the following form
\label{C3}
 \begin{equation}
\label{eq6}
|\Delta|\leq {8r_L}|J_\beta|{\kappa^{-2}(J_\beta,3)}\quad{\rm and}\quad||\mu_{\hat\beta}-\mu_\beta||^2\leq
{9r_L^2}|J_\beta| {\kappa^{-2}(J_\beta,3)}.
\end{equation}
\end{Corollary}
\begin{Corollary} Moreover, we have on ${\cal A}$
\label{C4}
 \begin{equation*}
\label{eq7}
||\Delta_J||\leq 3r_L|J_\beta|^{1/2}\kappa^{-2}(J_\beta,3)\quad{\rm and}
\quad |\Delta_J|\leq 3r_L|J_\beta|{\kappa^{-2}(J_\beta,3)}.
\end{equation*}
\end{Corollary}

\section{Concluding remarks}
In this paper we introduce the SOS algorithm for a linear model selection. The most computationally demanding part of the method is screening of predictors by the Lasso. Ordering and the greedy GIC can be computed using only two QR decompositions of $X_{0S_1}$. We give  non-asymptotic upper bounds on error probabilities of each step of the SOS in terms of the Lasso and GIC penalties (Theorem 1).  As a corollaries we obtain selection consistency for different ($n$, $p$) scenarios under conditions which are needed for screening consistency of the Lasso
(Corollaries 1-2). The SOS algorithm is an improvement of the new version of the thresholded Lasso \citep{Zhou09,Zhou10,GeerEtAl11} and turns out to be  a competitor for MCR, the latest quasiconvex penalized LS  \citep{ZhangT10,ZhangT13}. The condition on correlation of predictors  assumed there  seems to be stronger than ours but comparison of beta-min conditions requires inequalities between $\rho_-$ and $t^{-1/2}\kappa^2(T,3)$ and remains an  open problem.
For a traditional setting ($n >p$) we give Sanov-type bounds on error probabilities of  the OS algorithm (Theorem 2). Its surprising consequence is that the selection error of the greedy GIC is asymptotically not larger than of the exhaustive GIC (formula (\ref{greedGood})). Comparison of selection errors probabilities of the SOS and OS algorithms for $p < n$ requires further research. It is worth noticing that all results are proved for general form of  the Lasso  (formula (\ref{Lasso2})), which includes two versions of the estimator: algorithm  used in practice as well as its  formal version.


\appendix
\section*{Appendix A.}

\subsection*{A.1 Proofs for Section 3.}

\noindent
{\bf Proof of  Proposition  \ref{Prop1}.}
We have
\[  2\sigma^2KL(\tilde\beta_T^*||\tilde\beta_J)
 = 2\sigma^2\E_{\tilde\beta_T^*}\Bigg(\frac{||y-\tilde X_J\tilde\beta_J||^2-||y-\tilde X_T\tilde\beta_T^*||^2}{2\sigma^2}\Bigg)
= ||\tilde X_T\tilde\beta_T^* - \tilde X_J\tilde\beta_J||^2. \]
The last expression is symmetric with respect to $\tilde\beta_T^*$ and $\tilde\beta_J$, thus
$KL(\tilde\beta_T^*||\tilde\beta_J)=KL(\tilde\beta_J||\tilde\beta_T^*)$ and the second equality in (i) follows.
For the proof of the first equality in (i) observe that $\delta(T||J)=\min_{\tilde\beta_J}||\tilde X_T\tilde\beta_T^* - \tilde X_J\tilde\beta_J||^2$.
The equality in (ii) follows from (\ref{delta}), the inequality there follows from Rayleigh-Ritz theorem.
\hfill\BlackBox \\

\noindent
{\bf Proof of  Proposition  \ref{kappaProp}.}
 For the proof of (i) consider a  model $J$ and a  vector  $\nu$ such that $J \supseteq {\rm supp}(\nu)$ and  $|J|\leq (\lfloor c\rfloor +1)s$ and
$\kappa^2(\lfloor c\rfloor +1)s,0)=\nu^T\Sigma\nu/\nu^T\nu$. Sort coordinates of $\nu$ in nonincreasing order $\nu_{j_1}\geq \nu_{j_2}\ldots \geq \nu_{j_{(\lfloor c\rfloor +1)s}}$ and let $J_0=\{j_1,\ldots,j_s\}$. Then we have  $|J_0|=s$, $|\nu_{\bar J_0}|\leq
\lfloor c\rfloor|\nu_{ J_0}| \leq c|\nu_{ J_0}|$ and  $({\lfloor c\rfloor +1)}\nu_{ J_0}^T\nu_{J_0}\geq \nu^T\nu$. Thus
\[\kappa^2(s,c)\leq \frac{\nu^T\Sigma\nu}{\nu_{ J_0}^T\nu_{J_0}}\leq (\lfloor c\rfloor+1)\frac{\nu^T\Sigma\nu}{\nu^T\nu}= (\lfloor c\rfloor+1)\kappa^2((\lfloor c\rfloor+1)s,0) \]
and (i) follows.
\hfill\BlackBox \\

\noindent
{\bf Proof of Proposition \ref{kappaUnique}.}
Assume by contradiction that there are two different true models $T_1,T_2$
such that $T_i={\rm supp}(\beta_i)={\rm supp}(\theta_i)$ for some different $\beta_i=D\theta_i,~i=1,2$ and $\mu_0=X_0\theta_1=X_0\theta_2$.
It is enough to prove that assumptions imply $\gamma(T_1,1)\gamma(T_2,1)=0$, where $\gamma(J,c)=\inf\{||X_0\theta_J-X_0\theta_{\bar J}||, |\theta_J|=1,|\theta_{\bar J}|\leq c\}$ as in view of (\ref{kappaGamma}) and Schwarz inequality $\kappa(J,c)/\sqrt{|J|}\leq \gamma(J,c)$.
Define a vector $\theta$ with support equal to $T_1\cup T_2$ in such a way that $\theta_{T_1\cap T_2}=\theta_{T_1\cap T_2,1} - \theta_{T_1\cap T_2,2}$,
$\theta_{T_1\setminus T_2}=\theta_{T_1\setminus T_2,1}$ and $\theta_{T_2\setminus T_1}=\theta_{T_2\setminus T_1,2}$. As assumptions on $T_1$ and $T_2$ are symmetric we  may assume that
$|\theta_{T_1\setminus T_2}|\geq |\theta_{T_2\setminus T_1}|$ and let $\theta^{o}=\theta/|\theta_{T_1}|$. Then $|\theta^{o}_{T_1}|=1$ and
$|\theta^{o}_{\bar{T}_1}|=|\theta^{o}_{T_2\setminus T_1}| \leq 1$. Moreover, $X\theta^{o}_{{T}_1}=X\theta^{o}_{\bar{T}_1}$ which yields $\gamma(T_1,1)=0$.
\hfil\BlackBox \\

\noindent
{\bf Proof of Proposition \ref{kappaDelta}.}
To prove (i) observe that (\ref{deltaLambda}) and (\ref{kappa0}) imply for $j \in T$
\[  \kappa^2(T,3) \leq \kappa^2(T,0) \leq \theta_j^{*-2}\delta(T \parallel T\setminus\{j\}). \]
For (ii) we have
\begin{eqnarray*}
\kappa^2(t,3)/4 &\leq& \kappa^2(4t,0)=\min_{J:|J|\leq 4t}\lambda_{min}(\Sigma_J) \leq \min_{J: J\nsupseteq T,|J\cup T|\leq 4t}\lambda_{min}(\Sigma_J)\cr
 &=&  \min_{J: J\nsupseteq T,|J\cup T|\leq 4t}\lambda_{min}(\Sigma_{J\cup T}) \leq \theta_{min}^{*-2}\min_{J: J\nsupseteq T,|J\cup T|\leq 4t}\delta(T||J)\cr
&\leq& \theta_{min}^{*-2}\min_{J: J\nsupseteq T,|J\cup T|\leq 4t}\delta(T||J\setminus\{j\})= \theta_{min}^{*-2}\delta(T,4t_n),
\end{eqnarray*}
where the first inequality follows from the Proposition \ref{kappaProp} and the third from (\ref{deltaLambda}).
\hfill\BlackBox

\subsection*{A.2 Proofs for Section 6.}
We now proceed to prove Theorem \ref{SOI} and its corollaries. The following modified version of  Lemma 1 in \cite{BuneaEtAl07} holds.
\begin{Lemma}
(i) We have on ${\cal A}$ for  an arbitrary $\beta\in \R^p$ and $J=\{j:\beta_j\neq 0\}$
\begin{equation}
\label{eq1}
||\mu_0- \mu_{\hat\beta}||^2 +r_L|\Delta| \leq ||\mu_0-\mu_{\beta}||^2 +4r_L|\Delta_J|.
\end{equation}
(ii) Moreover, we have
 \begin{equation}
\label{eq2}
||\mu_0- \mu_{\hat\beta}||^2 \leq ||\mu_0-\mu_{\beta}||^2 +3r_L|\Delta_J|.
\end{equation}
\end{Lemma}

\noindent
{\bf Proof.}
It follows from (\ref{Lasso2}) that
\[||H_0(\varepsilon +\mu-X\hat\beta)||^2 +2r_L|D\hat\beta|\leq ||H_0(\varepsilon +\mu-X\beta)||^2 +2r_L|D\beta|\]
Equivalently, as $H_0$ is symmetric and idempotent, we get
\[||H_0(\mu-X\hat\beta)||^2 \leq  ||H_0(\mu-X\beta)||^2 +2\varepsilon^{T}H_0X(\hat \beta-\beta) +2r_L(|D\beta|-|D\hat\beta|).\]
Thus $||\mu_0-\mu_{\hat\beta}||^2 \leq  ||\mu_0-\mu_{\beta}||^2 +2{\varepsilon}^{T}X_0(\hat \theta-\theta) +2r_L(|\theta|-|\hat\theta|).$ §\\

\noindent
On ${\cal A}$ we have $|2\varepsilon^{T}X_0(\hat \theta-\theta)|\leq 2 \max_j|x_{0j}^T\varepsilon| |\hat \theta-\theta|\leq  r_L|\hat \theta-\theta|$ and whence on this set
\[||\mu_0-\mu_{\hat\beta}||^2 +r_L|\hat \theta-\theta| \leq  ||\mu_0-\mu_{\beta}||^2 +2r_{L} (|\hat \theta-\theta| +|\theta|-|\hat\theta|).\]
Note that for $j\not\in J$ $|\hat \theta_j-\theta_j| +|\theta_j|-|\hat\theta_j|=0$ and thus
\[||\mu_0-\mu_{\hat\beta}||^2 +r|\hat \theta-\theta|\leq ||\mu_0-\mu_{\beta}||^2 +2r_L( |\hat \theta_J-\theta_J| + |\theta_J|-|\hat \theta_J|).\]
Thus (i) follows from triangle inequality and (ii) from (i) in view of  $|\hat{\theta}_J-\theta_J| \leq |\hat\theta- \theta|$.
\hfill\BlackBox \\

\noindent
{\bf Proof of  Theorem \ref{SOI}.}
Proof of (i). Let $J=J_{\beta}$ and $\kappa=\kappa(J,3)$.  We consider two cases: (a) $|\Delta|>4|\Delta_J|$ and (b) $|\Delta|\leq 4|\Delta_J|$. In the case (a) it follows from (\ref{eq1}) that stronger inequality  $||\mu_0- \mu_{\hat\beta}|| \leq ||\mu_0-\mu_{\beta}||$ holds. When (b) is satisfied  we have $|\Delta_{\bar J}\leq 3 |\Delta_{J}|$  and it follows from the definition of $\kappa$ that
$\kappa^2||\Delta_J||^2\leq ||X_0\Delta||^2=||\mu_{\hat\beta}-\mu_{\beta}||^2$ and thus
\begin{equation}
\label{eq4}
||\Delta_J||\leq ||\mu_{\hat\beta}-\mu_{\beta}||\kappa^{-1}.
\end{equation}
Using (\ref{eq4}) and Jensen inequality we get
 \begin{equation}
\label{eq5}
|\Delta_J|\leq |J|^{1/2}||\mu_{\hat\beta}-\mu_{\beta}||\kappa^{-1}.
\end{equation}
It follows  now from  (\ref{eq2}), (\ref{eq5}) and triangle inequality that
\[||\mu_0- \mu_{\hat\beta}||^2 \leq ||\mu_0- \mu_{\beta}||^2 +3r_L|J|^{1/2}\kappa^{-1}(||\mu_0- \mu_{\hat\beta}|| +  ||\mu_0- \mu_{\beta}||)\]
and whence
\[(||\mu_0+ \mu_{\hat\beta}|| +  ||\mu_0- \mu_{\beta}||)(||\mu_0- \mu_{\hat\beta}|| - ||\mu_0- \mu_{\beta}||)\leq 3r_L|J|^{1/2}\kappa^{-1}(||\mu_0- \mu_{\hat\beta}|| +  ||\mu_0- \mu_{\beta}||)\]
from which the conclusion follows.\\

Proof of (ii). Define  $m=||\mu_0- \mu_{\beta}||$, $\hat m=||\mu_0- \mu_{\hat\beta}||$  and $c=2r_L|J|^{1/2}\kappa^{-1}$. Using (\ref{eq1}), (\ref{eq5}) which holds provided $|\Delta|\leq 4|\Delta_J|$, and triangle inequality we get
\[\hat m^2 +r_L|\Delta|\leq m^2 + 2c(\hat m +m) \leq2 m^2 +c^2 + \hat m^2 +c^2, \]
from which the desired bound follows.
\hfill\BlackBox \\

\noindent
{\bf Proof of  Corollary \ref{C3}.}
Put $c=3r|J|^{1/2}\kappa^{-1}$ and $m=||\mu_0- \mu_{\beta}||$. Squaring both sides of (\ref{eq3}) we get
\[ ||\mu_0- \mu_{\hat\beta}||^2 \leq (m+c)^2=m^2 +c^2 +2\frac{c}{\sqrt{a}}\sqrt{a}m\leq (1+a)(m^2+c^2/a),\]
from which the conclusion immediately follows.
\hfill\BlackBox \\

\noindent
{\bf Proof of  Corollary \ref{C4}.}
The proof follows from  inequality (\ref{eq4}), (\ref{eq5}) and the second inequality in Corollary \ref{C3}.
\hfill\BlackBox

\subsection*{A.3 Proofs for Section 4.}
The next lemma states bounds on upper tail of $\chi_k^2$ distribution

\begin{Lemma}
\label{L2}
Let $W_k$ denote variable having $\chi_k^2$ distribution.(i) (Gordon, 1941 and Mill, 1926) We have for $k=1$ and $x>0$
\begin{equation}
\label{boundschi1}
w_{xk}l_{xk}\leq P( W_k\geq x)\leq w_{xk},
\end{equation}
where $w_{xk}=e^{-x/2}(\frac{x}{2})^{k/2-1}\Gamma^{-1}(\frac{k}{2})$ and $l_{xk}=\frac{x}{x-k+2}$.\\
(ii) \citep{InglotLedwina06} Let $k>1$ and $x>k-2$. Then
\begin{equation}
\label{boundschi2}
w_{xk}\leq P( W_k\geq x)\leq w_{xk}l_{xk}.
\end{equation}
\end{Lemma}

\noindent
{\bf Proof.}
We provide  the unified  reasoning  for both cases. For $x>0$ and $k\in\Z$ let $I_k(x)=\int_x^\infty t^{(k/2)-1}e^{-t/2}\,dt$. Integration by parts yields
\begin{equation}
\label{A1}
I_k(x)=2x^{(k/2)-1}e^{-x/2} +(k-2)I_{k-2}(x).
\end{equation}
It is easy to see that the following inequalities hold  for $x>0$ and $k\in\Z$
\begin{equation}
\label{A2}
0\leq I_{k-2}(x)\leq I_k(x)/x.
\end{equation}
We treat cases $k=1$ and $k>1$ separately, as $k=1$ is the only integer for which the second term on the RHS of (\ref{A1})
is negative. Dividing both sides of (\ref{A1})  by $2^{k/2}\Gamma(k/2)$, noting that the LHS is then $p(x|\chi_k^2)$ and using
(\ref{A2})  we have for $k=1$ and $x>0$
\begin{equation*}
\label{A3}
p(x|\chi_1^2) \leq e^{-x/2}\Big(\frac{x}{2}\Big)^{-1/2}\Gamma^{-1}\Big(\frac{1}{2}\Big)
\end{equation*}
and
\begin{equation*}
\label{A4}
p(x|\chi_1^2) \geq e^{-x/2}\Big(\frac{x}{2}\Big)^{-1/2}\Gamma^{-1}\Big(\frac{1}{2}\Big)\Big(1-\frac{1}{1+x}\Big),
\end{equation*}
which proves (\ref{boundschi1}). Analogously for $k=2,3,\ldots$ we obtain  from (\ref{A1}) inequalities proved by \cite{InglotLedwina06}
\begin{equation*}
\label{A5}
p(x|\chi_k^2) \leq e^{-x/2}\Big(\frac{x}{2}\Big)^{k/2-1}\Gamma^{-1}\Big(\frac{k}{2}\Big)\Big(1+\frac{k-2}{x-k+2}\Big)
\end{equation*}
 for $x>k-2$, and for $x>0$
\begin{equation*}
\label{A6}
p(x|\chi_k^2) \geq e^{-x/2}\Big(\frac{x}{2}\Big)^{k/2-1}\Gamma^{-1}\Big(\frac{k}{2}\Big),
\end{equation*}
which proves (\ref{boundschi2}).
\hfill\BlackBox \\

Now we state the main lemma from which Theorems \ref{Th1} and \ref{Th2}  follow. Let us recall that
$c_1=(6+4\sqrt{2})^{-1}$ and $c_2=(3+6\sqrt{2})^{-1}$. Define ${\cal T}_n^o={\cal T}_n \setminus \{T\}$ and
observe that for OS algorithm we have $P( S_1\not\in {\cal T}_n)=0$ and  as $p\geq t+1$, ${\cal T}_n={\cal T}_n^o=\{F\}$, so $|{\cal T}_n^o|=1$.
\begin{Lemma}
\label{Lem1}
(T1) If $r_L^2\leq c_1^2t^{-1}\kappa^4\theta_{min}^{*2}$, then
\begin{equation*}
\label{T1}
P( S_1\not\in {\cal T}_n) \leq p \exp \bigg(-\frac{r_L^2}{8\sigma^2}\bigg) \bigg(\frac{\pi r_L^2}{8\sigma^2}\bigg)^{-1/2}.
\end{equation*}
(T2) If $s\leq n$, then
\begin{equation*}
\label{T2}
P( S_1\in {\cal T}_n, \hat O\not \in O_{S_1} ) \leq \frac{3}{2}|{\cal T}_n^o|t(s-t) \exp\bigg(-\frac{c_2\delta_s}{\sigma^2}\bigg)\bigg(\frac{\pi c_2\delta_s}{\sigma^2}\bigg)^{-1/2}.
\end{equation*}
(T3) If for some $a\in (0,1)~~r \leq at^{-1}\delta_t$, then
\begin{equation*}
\label{T3}
P( S_1\in {\cal T}_n, \hat O\in O_{S_1}, |\hat T|<t )\leq \frac{t}{2} \exp\bigg(-\frac{(1-a)^2\delta_t}{8\sigma^2}\bigg)\bigg(\frac{\pi(1-a)^2\delta_t}{8\sigma^2}\bigg)^{-1/2}.
\end{equation*}
(T4) Assume that  $r/\sigma^2\geq 2$ and  $(r/\sigma^2) -\log (r/\sigma^2) \geq 2\log p$. Then
\begin{equation*}
\label{T4}
P( S_1\in {\cal T}_n, \hat O \in O_{S_1}, |\hat T |>t )\leq
(p-t)(s-t) \exp \bigg(-\frac{r}{2\sigma^2}\bigg)\bigg(\frac{\pi r}{2\sigma^2}\bigg).
\end{equation*}
\end{Lemma}

\noindent
{\bf Proof.}
Observe that we may assume that $t>0$ in proofs of $(T2)-(T3)$  as  for $t=0$  probabilities appearing in those parts are 0 and the conclusions are trivially satisfied. \\

Proof of (T1).
It follows from  (\ref{boundA}) or equivalently from Lemma \ref{L2} that it is enough to prove that $\{S_1 \in {\cal T}_n \} \supseteq {\cal A}$ that is that on ${\cal A}$ we have
\begin{equation}
\label{boundsTL}
T\subseteq S_1\quad\quad{\rm and}  \quad\quad |S_1|\leq t +\lfloor\sqrt{t}\kappa^{-2}\rfloor.
\end{equation}
For parametric models we have $|\Delta|\leq 4|\Delta_T|$ or equivalently $4|\Delta_{\bar T}|\leq 3|\Delta|$, which together with the first  part of (\ref{eq6})  yields $|\Delta_{\bar T}|\leq 6rt\kappa^{-2}$. Thus $|S_0\setminus T|\leq |\Delta_{\bar T}|/a_0\leq t\kappa^{-2}$, $|S_0|\leq t(1+\kappa^{-2})$ and
$a_1\leq 6r_L\sqrt{t(1+\kappa^{-2})}$. Using this and (\ref{eq6}) we have $||\Delta_T|| +a_1\leq \theta^*_{min}$ or
\begin{equation}
\label{eq9}
||\Delta_T||^2\leq (\theta^*_{min}-a_1)^2
\end{equation}
Indeed, from Corollary \ref{C4}, the fact  that $\kappa\leq 1$ and the assumption of the lemma, respectively,  we have
\begin{eqnarray*}
&||\Delta_T|| +a_1 \leq 3r_Lt^{1/2}\kappa^{-2} +6r_L\sqrt{t(1+\kappa^{-2})} = 3r_Lt^{1/2}\kappa^{-2}(1+2\sqrt{\kappa^4+\kappa^2})\cr
&\leq 3(1+2\sqrt{2})r_Lt^{1/2}\kappa^{-2} = c_1^{-1}r_Lt^{1/2}\kappa^{-2} \leq \theta^*_{min}.\cr
\end{eqnarray*}
Evidently, $|T\setminus S_1|(\theta^*_{min}-a_1)^2 < ||\Delta_T||^2\leq (\theta^*_{min}-a_1)^2$ and thus in view of (\ref{eq9}) we have $T\subseteq S_1$ on ${\cal A}$. But $S_1\subseteq S_0$, thus $|S_0|\geq t$ and $a_1\geq 6r_Lt^{1/2}$. Thus using (\ref{eq6}) again, we have
$§ |S_1\setminus T|\leq |\Delta_{\bar T}|/a_1\leq  t^{1/2}\kappa^{-2}.$
Hence  $|S_1\setminus T|\leq \lfloor t^{1/2}\kappa^{-2}\rfloor$ and  we obtain (\ref{boundsTL}). \\

Proof of (T2). Let for $J_1\in {\cal S}_n \setminus {\cal T}_n$ and $J_2\in {\cal T}_n$   $W_{J_1J_2}=\varepsilon^T(\tilde H_{J_1} -\tilde H_{J_1\cap J_2})\varepsilon$,
 $\sigma^2W_{J_2J_1}=\varepsilon^T(\tilde H_{J_2} -\tilde H_{J_1\cap J_2})\varepsilon$ and $\sigma Z_{J_1}= \tilde\beta_T^{*T}\tilde X_T^T(I-\tilde H_{J_1})\varepsilon/\sqrt{\delta_{J_1}}$, where $\delta_{J_1}=\delta(T\parallel J_1)$.
Then we have that $W_{J_1J_2}\sim \chi_d^2$, where $d\leq |J_1\setminus J_2|$, $W_{J_2J_1}\geq 0$  and $Z_{J_1}\sim N(0,1)$. We will use a popular decomposition  of a difference between sums of squared residuals
 \begin{eqnarray*}
 \label{RSS}
 R_{J_1}- R_{J_2} &=&\tilde\beta_T^{*T}\tilde X_T^T(I-\tilde H_{J_1})\tilde X_T\tilde\beta_T^* + 2\tilde\beta_T^{*T}\tilde X_T^T(I-\tilde H_{J_1})\varepsilon \cr
 &+&\varepsilon^T(I-\tilde H_{J_1})\varepsilon - \varepsilon^T(I-\tilde H_{J_2})\varepsilon \cr
 &=&\delta_{J_1} +2\sqrt{\delta_{J_1}}\sigma Z_{J_1} -\sigma^2 W_{J_1J_2} +  \sigma^2W_{J_2J_1} \cr
 &\geq& \delta_{J_1}\Big( 1 +\frac{2\sigma Z_{J_1}}{\sqrt{\delta_{J_1}}} -\frac{ \sigma^2 W_{J_1J_2}}{\delta_{J_1}}\Big).
 \end{eqnarray*}
For fixed $S\in {\cal T}_n^o$  let  $\bar j=S\setminus \{j\}$. Then we have
 \begin{eqnarray*}
 \label{RSS1}
\{ S_1\in {\cal T}_n^o, \hat O\not \in O_{S_1} \} &\subseteq& \bigcup_{S\in {\cal T}_n^o}\bigcup_{j_1\in T}\bigcup_{j_2\in S\setminus T}\{ R_{\bar j_1}<  R_{\bar j_2}\}\cr
&\subseteq&  \bigcup_{S\in {\cal T}_n^o}\bigcup_{j_1\in T}\bigcup_{j_2\in S\setminus T}\{ -\frac{2\sigma Z_{\bar j_1}}{\sqrt{\delta_{\bar  j_1}}}  +\frac{\sigma^2 W_{\bar j_1\bar j_2}}{\delta_{\bar j_1}}  >1\},
\end{eqnarray*}
where $Z_{\bar j_1}\sim N(0,1)$ and $W_{\bar j_1\bar j_2}\sim \chi_d^2$, with $d\leq 1$.
Thus it follows that for $W=Z^2$ denoting  r.v. with $\chi^2_1$ distribution, we get
\begin{eqnarray*}
 \label{RSS2}
P( S_1\in {\cal T}_n^o, \hat O\not \in O_{S_1} )&\leq&  \sum_{S\in {\cal T}_n^o}\sum_{j_1\in T}\sum_{j_2\in S\setminus T}P(-\frac{2\sigma Z_{\bar j_1}}{\sqrt{\delta_{\bar  j_1}}}  +\frac{ \sigma^2 W_{\bar j_1\bar j_2}}{\delta_{\bar j_1}}\geq 1)\cr
&\leq&  \sum_{S\in {\cal T}_n^o}\sum_{j_1\in T}\sum_{j_2\in S\setminus T}\Big(P(-\frac{2\sigma Z_{\bar j_1}}{\sqrt{\delta_{\bar  j_1}}}\geq c) +P(\frac{\sigma^2 W_{\bar j_1\bar j_2}}{\delta_{\bar j_1}}\geq 1-c)\Big)\cr
&\leq& |{\cal T}_n^o|t(s -t)\Big(\frac{1}{2}P(Z^2\geq \frac{c^2\delta_s}{4\sigma^2}) +P( W\geq\frac{(1-c)\delta_s}{\sigma^2}\Big),
\end{eqnarray*}
where $j_1\in T$ and $j_2\in S\setminus T$ are fixed and we used $\delta_{\bar  j_1}\geq \delta_s$. Choosing $c$ such that $c^2/4= 1-c$ that is $c=1-2c_2$ in view of Lemma \ref{L2} we get the desired bound.\\

Proof of (T3). Reasoning as previously we have
\[ \{S_1\in {\cal T}_n, \hat O\in O_{S_1}, |\hat T|<t\}\subseteq\bigcup_{S\subset T}
\{R_S + r|S|\leq R_T + r|T|\}\subseteq\bigcup_{j\in T}\{R_{\bar j} \leq  R_T +rt\}.\]
Thus in view of Lemma \ref{L2} and the assumption $rt <a\delta_t$ we obtain
\begin{eqnarray*}
 \label{T31}
P( S_1\in {\cal T}_n, \hat O\in O_{S_1}, |\hat T|<t ) &\leq& \sum_{j\in T} P( R_{\bar j} \leq R_T +rt) \cr
&\leq& \sum_{j\in T} P\Big(-{2\sigma Z_{\bar j}}\geq \sqrt{\delta_{\bar  j}}\Big(1-\frac{rt}{\delta_{\bar j}}\Big)\Big)\cr
&\leq& t P\Big(-2\sigma Z\geq {\sqrt{\delta_{t}}}\Big(1-\frac{rt}{\delta_{t}}\Big)\Big) \cr
&=&\frac{1}{2}t P\Big(W \geq \frac{1}{4\sigma^2}\delta_t\Big(1-\frac{rt}{\delta_t}\Big)^2\Big)\cr
&\leq& \frac{t}{2}\exp \bigg(-\frac{(1-a)^2\delta_t}{8\sigma^2}\bigg) \bigg(\frac{\pi(1-a)^2\delta_t}{8\sigma^2}\bigg)^{-1/2}.
 \end{eqnarray*}

Proof of (T4). Observe first that for $m>0$
 \begin{eqnarray*}
 & & P(S_1\in {\cal T}_n, \hat O \in O_{S_1}, |\hat T |=t+m )\cr & &\leq P(R_{T\cup \{j_1,\ldots,j_m\}} +(t +m)r \leq  R_T +tr \,\, {\rm for \,\, some}\,\, j_1,\ldots,j_m\in  F\setminus T)
 \cr & & \leq {p-t\choose m}P(\sigma^2W_m\geq mr)\leq \frac{(p-t)^m}{m!}P(\sigma^2W_m\geq mr)=B_m,
 \end{eqnarray*}
where $W_m\sim \chi^2_m$. This follows  since for any fixed $J= T\cup\{j_1,\ldots,j_m\}$ $R_T -R_J\sim \sigma^2\chi_d^2$, where $d\leq m$ and $W_d\leq W_m$ in stochastic order. We will show that under conditions given in (T4) $B_m\geq B_{m+1}$ for any $m=1,2,\ldots$ thus yielding
\[ P(S_1\in {\cal T}_n, \hat O \in O_{S_1}, |\hat T |\geq t+m ) \leq (s-t) B_m,\]
which for $m=1$ coincides with the desired inequality. Let $Q_m=B_m/B_{m+1}$, $\bar r=r/\sigma^2$ and observe that for $m>1$ we have in view of  (\ref{boundschi2}) (note that $m\bar r\geq m-2$ as $\bar r\geq 2$)
\[ Q_m\geq  \frac{m+1}{p}e^{\bar r/2}\big(\frac{m}{m+1}\Big)^{m/2-1}\frac{1}{\big((m+1)\bar r/2\big)^{1/2}}\frac{\Gamma((m+1)/2)}{\Gamma(m/2)}\frac{(m+1)\bar r-m +1}{(m+1)\bar r}.\]
Using the inequality for gamma functions \citep[cf formula 2.2 in][]{Laforgia84}
\[ \Gamma \Big(\frac{m+1}{2} \Big) \Big/ \Gamma \Big(\frac{m}{2} \Big)\geq \Big(\frac{m-1/2}{2}\Big)^{1/2}\]
we have that
\[  Q_m \geq \exp\Big\{ \frac{\bar r}{2} - \frac{1}{2}\log \bar r -\log p\Big\}f_1(m,\bar r),\]
where
\[ f_1(m,\bar r)=\Big(\frac{m}{m+1}\Big)^{m/2-1}(m+1)^{1/2}2^{1/2}\Big(\frac{m-1/2}{2}\Big)^{1/2}\frac{(m+1)\bar r-m+1}{(m+1)\bar r}.\]
Thus in order to show that $Q_m>1$ for $m>1$ in view of assumptions it is enough to show that $f_1(m,\bar r)>1$. As $f(m,\cdot)$ is increasing, it suffices to check that $f_1(m,2)>1$.
 Let $f_2(m)=(\frac{m-1/2}{m+1})^{(m-1)/2}(\frac{m+3}{2})$. We have $f_1(m,2)>f_2(m)$ and $f_2(2)>1$ thus it is enough to show that $f_2$ is increasing. Let
 \[ f_3(m)=\log(2f_2(m))=\frac{m-1}{2}\log \frac{m-1/2}{m+1}+ \log(m+3).\]
 We have that
 \begin{eqnarray*}
  f_3'(m)&=&\frac{1}{2}\log \frac{m-1/2}{m+1} + \frac{m-1}{2}\frac{m+1}{(m-1/2)}\frac{3}{2(m+1)^2}+\frac{1}{m+3}\cr
  &\geq& \frac{1}{2}\frac{-3}{-3 +2(m+1)} + \frac{3(m-1)}{4(m-1/2)(m+1)} +\frac{1}{m+3},
  \end{eqnarray*}
  where the last inequality follows from $\log(1+x)>x/(1+x)$ for $x>-1$. As $1/(m+3)\geq 3/(-6+2(m+1))$ it follows that $f_3'>0$ which implies that $f_3$ and thus $f_2$ is increasing.
\hfill\BlackBox \\

\noindent
{\bf Proof of Theorem \ref{Th1}.}
The result readily follows from  Lemma \ref{Lem1}. For (T1) we observe that
\[ -\frac{r_L^2}{8\sigma^2} +\log p\leq -\frac{(1-a)r_L^2}{8\sigma^2}\]
is equivalent to $8\sigma^2a^{-1}\log p\leq r_L^2$. Similar reasoning yields (T4).
Consider  derivation of (T2).  From the bound
\[ |{\cal T}_n^o|=|{\cal T}_n|-1=\sum_{k=1}^{s-t}{p-t\choose k}\leq (p-t)+\ldots+\frac{(p-t)^{s-t}}{(s-t)!}\leq \frac{(p-t)^{s-t}}{(s-t)!}(s-t)\]
it follows that $ |{\cal T}_n^o|t(s-t)\leq (p-t)^{s-t}t(s-t)\leq  p^{s-t}t(s-t).$
Thus the bound in  (T2) will follow from
$ -c_2\delta_n/\sigma^2 +(s-t)\log p + \log(s-t) + \log t\leq -c_2(1-a)\delta_s/\sigma^2 $
which is implied by $(s-t+2)\log p \leq c_2a\delta_s/\sigma^2$. For (T3) we observe that
\[ -\frac{(1-a)^2\delta_t}{8\sigma^2} +\log t\leq -\frac{(1-a)^3\delta_t}{8\sigma^2}\]
is equivalent to $8\sigma^2\log t\leq (1-a)^2 a\delta_t$.
\hfill\BlackBox \\

\noindent
{\bf Proof of  Corollary \ref{cor1}.}
We proceed by showing that assumptions (i) and (ii)  imply all assumptions of Theorem \ref{Th1}.
We first note that (i) with the assumption $r_L^2=4r$ is stronger than the assumption in Theorem \ref{Th1} (T1). Next, observe that condition
\begin{equation*}
 \label{cor1e1}
4a^{-1}\sigma^2\log p\leq (4c_2/3) t^{-1/2}\kappa^{2}\delta_s
\end{equation*}
is stronger than the assumption in Theorem \ref{Th1} (T2). Indeed, as $\kappa\leq 1 \leq t$  we have
\[ s-t+2=\lfloor t^{1/2}\kappa^{-2}\rfloor +2\leq  t^{1/2}\kappa^{-2} +2\leq 3 t^{1/2}\kappa^{-2}.\]
Obviously, left inequalities in (i) and (ii) imply (\ref{cor1e1}). Moreover, the assumption of Theorem \ref{Th1} (T4) is satisfied.
Furthermore, from the first $\kappa-\delta$ inequality (\ref{betadelta1}) and assumption $a \in (0,1-c_1)$ we obtain that (i) is stronger than both conditions in Theorem \ref{Th1} (T3).

In order to justify the conclusion, in view of the fact that $e^{-(1-a)x}(\pi x)^{-1/2}$ is decreasing function of $x>0$, it is enough to show that the expressions in the exponents of the bounds (\ref{L22}) and (\ref{L23}) are larger than $r/(2\sigma^2)$ that is a value in  the exponents of the bounds (\ref{L21}) and (\ref{L24}) . In the case of (\ref{L22}) the condition is equivalent to $r \leq 2c_2\delta_s$, which is implied
by (ii). In the case of (\ref{L23}) the ensuing inequality is implied by $r \leq ((1-a)^2/4)\kappa^2\theta_{min}^{*2}$ which in turn is implied by (i) as $a \in (0,1-c_1)$.
\hfill\BlackBox \\

\noindent
{\bf Proof of Theorem \ref{Th2}.}
Let us recall that for OS algorithm we have
$P( S_1\not\in {\cal T}_n)=0$ and $|{\cal T}_n^o|=1$, so the results follow from  Lemma \ref{Lem1} analogously to Theorem \ref{Th1}.
\hfill\BlackBox \\

\noindent
{\bf Proof of  Corollary \ref{cor3}.}
We proceed as in  the proof of  Corollary \ref{cor1}. The following condition
\begin{equation}
 \label{cor3e1}
4a^{-1}\sigma^2\log p\leq 2c_2 \delta_s.
\end{equation}
is stronger than the assumption in Theorem \ref{Th2}. The assumption imply (\ref{cor3e1}) and the assumption of (T4).
Furthermore, from the first $\kappa-\delta$ inequality (\ref{betadelta1}) and assumption $a \in (0,2c_2)$ we obtain that the assumption is stronger than both conditions in (T3).

Next we show that the powers in the exponents of the bounds (\ref{L22}) and (\ref{L23}) are larger than $r/(2\sigma^2)$. In the case of (\ref{L22}) the condition is equivalent to $r \leq 2c_2\delta_s$ which is implied by the assumption. In the case of (\ref{L23}) the ensuing inequality is implied by $r \leq ((1-a)^2/4)\delta_t$,  which is implied by $r \leq at^{-1}\delta_t$ because for $a \in (0,1)$ a condition $a \leq (1-a)^2/4$ is equivalent to $a \in (0,2c_2)$.
\hfill\BlackBox \\

\bibliography{PokarowskiMielniczukJMLR13.bib}

\begin{thebibliography}{31}
\providecommand{\natexlab}[1]{#1}
\providecommand{\url}[1]{\texttt{#1}}
\expandafter\ifx\csname urlstyle\endcsname\relax
  \providecommand{\doi}[1]{doi: #1}\else
  \providecommand{\doi}{doi: \begingroup \urlstyle{rm}\Url}\fi

\bibitem[Bickel et~al.(2009)Bickel, Ritov, and Tsybakov]{BickelEtAl09}
P.~Bickel, Y.~Ritov, and A.~Tsybakov.
\newblock Simultaneous analysis of {L}asso and {D}antzig selector.
\newblock \emph{Annals of Statistics}, 37:\penalty0 1705--1732, 2009.

\bibitem[B\"uhlmann and van~de Geer(2011)]{BuhlmannGeer11}
P.~B\"uhlmann and S.~van~de Geer.
\newblock \emph{Statistics for High-dimensional Data}.
\newblock Springer, New York, 2011.

\bibitem[Bunea et~al.(2007)Bunea, Tsybakov, and Wegkamp]{BuneaEtAl07}
F.~Bunea, A.~Tsybakov, and M.~Wegkamp.
\newblock Sparsity oracle inequalities for the {L}asso.
\newblock \emph{Electronic Journal of Statistics}, 37:\penalty0 169--194, 2007.

\bibitem[Casella et~al.(2009)Casella, Giron, Martinez, and
  Moreno]{CasellaEtAl09}
G.~Casella, F.~Giron, M.~Martinez, and E.~Moreno.
\newblock Consistency of bayesian procedures for variable selection.
\newblock \emph{Annals of Statistics}, 37:\penalty0 1207--1228, 2009.

\bibitem[Chen and Chen(2008)]{ChenChen08}
J.~Chen and Z.~Chen.
\newblock Extended bayesian information criterion for model selection with
  large model spaces.
\newblock \emph{Biometrika}, 95:\penalty0 759--771, 2008.

\bibitem[Cover and Thomas(2006)]{CoverThomas06}
T.~Cover and J.~Thomas.
\newblock \emph{Elements of Information Theory}.
\newblock Wiley, New York, 2006.

\bibitem[Fan and Li(2001)]{FanLi01}
J.~Fan and R.~Li.
\newblock Variable selection via nonconcave penalized likelihood and its oracle
  properties.
\newblock \emph{Journal of the American Statistical Association}, 96:\penalty0
  1348--1360, 2001.

\bibitem[Huang and Zhang(2012)]{HuangZhang12}
J.~Huang and C.H. Zhang.
\newblock Estimation and selection via absolute penalized convex minimization
  and its multistage adaptive applications.
\newblock \emph{Journal of Machine Learning Research}, 13:\penalty0 1839--1864,
  2012.

\bibitem[Huang et~al.(2008)Huang, Ma, and Zhang]{HuangEtAl08}
J.~Huang, S.~Ma, and C.H. Zhang.
\newblock Adaptive {L}asso for sparse high-dimensional regression models.
\newblock \emph{Statistica Sinica}, 18:\penalty0 1603--1618, 2008.

\bibitem[Inglot and Ledwina(2006)]{InglotLedwina06}
T.~Inglot and T.~Ledwina.
\newblock Asymptotic optymality of new adaptive test in regression model.
\newblock \emph{Annales de l'Institut Henri Poincare. Probability and
  Statistics}, 42:\penalty0 579--590, 2006.

\bibitem[Laforgia(1984)]{Laforgia84}
A.~Laforgia.
\newblock Further inequalities for the gamma function.
\newblock \emph{Mathematics of Computation}, 42:\penalty0 597--600, 1984.

\bibitem[Luo and Chen(2013)]{LuoChen13}
S.~Luo and Z.~Chen.
\newblock Extended {BIC} for linear regression models with diverging number of
  relevant features and high or ultra-high feature spaces.
\newblock \emph{Journal of Statistical Planning and Inference}, 143:\penalty0
  494--504, 2013.

\bibitem[Meinshausen and B\"uhlmann(2006)]{MeinshausenBuhlmann06}
N.~Meinshausen and P.~B\"uhlmann.
\newblock High dimensional graphs and variable selection with the {L}asso.
\newblock \emph{Annals of Statistics}, 34:\penalty0 1436--1462, 2006.

\bibitem[Meinshausen and Yu(2009)]{MeinshausenYu09}
N.~Meinshausen and B.~Yu.
\newblock Lasso-type recovery of sparse representations for high-dimensional
  data.
\newblock \emph{Annals of Statistics}, 37:\penalty0 246--270, 2009.

\bibitem[P\"otscher and Schneider(2011)]{PotscherSchneider11}
B.M. P\"otscher and U.~Schneider.
\newblock Distributional results for thresholding estimators in
  high-dimensional gaussian regression models.
\newblock \emph{Electronic Journal of Statistics}, 5:\penalty0 1876--1934,
  2011.

\bibitem[Rao and Wu(1989)]{RaoWu89}
C.R. Rao and Y.~Wu.
\newblock Strongly consistent procedure for model selection in a regression
  problem.
\newblock \emph{Biometrika}, 76:\penalty0 369--374, 1989.

\bibitem[Shao(1998)]{Shao98}
J.~Shao.
\newblock Convergence rates of the generalized information criterion.
\newblock \emph{Journal of Nonparametric Statistics}, 9:\penalty0 217--225,
  1998.

\bibitem[Tibshirani(1996)]{Tibshirani96}
R.~Tibshirani.
\newblock Regression shrinkage and selection via the {L}asso.
\newblock \emph{Journal of the Royal Statistical Society Series B},
  58:\penalty0 267--288, 1996.

\bibitem[Tibshirani(2011)]{Tibshirani11}
R.~Tibshirani.
\newblock Regression shrinkage and selection via the {L}asso: a retrospective.
\newblock \emph{Journal of the Royal Statistical Society Series B},
  73:\penalty0 273--282, 2011.

\bibitem[van~de Geer and B\"uhlmann(2009)]{GeerBuhlmann09}
S.~van~de Geer and P.~B\"uhlmann.
\newblock Consistency of bayesian procedures for variable selection.
\newblock \emph{Electronic Journal of Statistics}, 3:\penalty0 1360--1392,
  2009.

\bibitem[van~de Geer et~al.(2011)van~de Geer, B{\"u}hlmann, and
  Zhou]{GeerEtAl11}
S.~van~de Geer, P.~B{\"u}hlmann, and S.~Zhou.
\newblock The adaptive and the thresholded {L}asso for potentially misspecified
  models (and a lower bound for the {L}asso).
\newblock \emph{Electronic Journal of Statistics}, 5:\penalty0 688--749, 2011.

\bibitem[Zhang(2010{\natexlab{a}})]{ZhangCH10}
C.H. Zhang.
\newblock Nearly unbiased variable selection under minimax concave penalty.
\newblock \emph{Annals of Statistics}, 38:\penalty0 894--942,
  2010{\natexlab{a}}.

\bibitem[Zhang and Zhang(2012)]{ZhangZhang12}
C.H. Zhang and T.~Zhang.
\newblock A general theory of concave regularization for high-dimensional
  sparse estimation problems.
\newblock \emph{Statistical Science}, 27:\penalty0 576--593, 2012.

\bibitem[Zhang(2010{\natexlab{b}})]{ZhangT10}
T.~Zhang.
\newblock Analysis of multi-stage convex relaxation for sparse regularization.
\newblock \emph{Journal of Machine Learning Research}, 11:\penalty0 1081--1107,
  2010{\natexlab{b}}.

\bibitem[Zhang(2013)]{ZhangT13}
T.~Zhang.
\newblock Multistage convex relaxation for feature selection.
\newblock \emph{submited to Bernoulli}, 2013.

\bibitem[Zhao and Yu(2006)]{ZhaoYu06}
P.~Zhao and B.~Yu.
\newblock On model selection consistency of {L}asso.
\newblock \emph{Journal of Machine Learning Research}, 7:\penalty0 2541--2563,
  2006.

\bibitem[Zheng and Loh(1995)]{ZhengLoh95}
H.~Zheng and W.~Loh.
\newblock Consistent variable selection in linear models.
\newblock \emph{Journal of the American Statistical Association}, 90:\penalty0
  151--156, 1995.

\bibitem[Zhou(2009)]{Zhou09}
S.~Zhou.
\newblock Thresholding procedures for high dimensional variable selection and
  statistical estimation.
\newblock In \emph{NIPS}, pages 2304--2312, 2009.

\bibitem[Zhou(2010)]{Zhou10}
S.~Zhou.
\newblock Thresholded {L}asso for high dimensional variable selection and
  statistical estimation.
\newblock \emph{ArXiv}, 2010.

\bibitem[Zou(2006)]{Zou06}
H.~Zou.
\newblock The adaptive lasso and its oracle properties.
\newblock \emph{Journal of the American Statistical Association}, 101:\penalty0
  1418--1429, 2006.

\bibitem[Zou and Li(2008)]{ZouLi08}
H.~Zou and R.~Li.
\newblock One-step sparse estimates in nonconcave penalized likelihood models.
\newblock \emph{Annals of Statistics}, 36:\penalty0 1509--1533, 2008.

\end{thebibliography}
\end{document}